\documentclass[sigconf]{acmart}

\AtBeginDocument{%
  \providecommand\BibTeX{{%
    \normalfont B\kern-0.5em{\scshape i\kern-0.25em b}\kern-0.8em\TeX}}}

\copyrightyear{2024}
\acmYear{2024}
\setcopyright{acmlicensed}\acmConference[KDD'24]{}{Aug 25--29, 2024}{Barcelona, Spain}
\acmPrice{15.00}
\acmDOI{10.1145/3437963.3441810}
\acmISBN{978-1-4503-8297-7/21/03}

\def\ie{\textit{i.e.,~}}
\def\E{\mathbb{E}}

\usepackage[ruled,linesnumbered]{algorithm2e}
\usepackage{booktabs}  
\usepackage{threeparttable} 
\usepackage{multirow}

\usepackage{amssymb}
\usepackage{bm}
\usepackage{subcaption} 
\usepackage{threeparttable}
\usepackage[normalem]{ulem}
\useunder{\uline}{\ul}{}

\newcommand{\vpara}[1]{\vspace{0.04in}\noindent\textbf{#1}\xspace}
\begin{document}

% \title{Efficient Graph Contrastive Learning with Ranking}
% \title{Graph Ranking Contrastive Learning: A  Simple but Efficient Method}
\title{Graph Ranking Contrastive Learning: A Extremely Simple yet Efficient Method}

\author{Yulan Hu}
\email{huyulan@ruc.edu.cn}
\affiliation{
  \institution{Renmin University of China, Kuaishou Technology}
   \country{}
}

\authornote{The three authors contributed equally to this research.}

\author{Sheng Ouyang}
\email{ouyangsheng@ruc.edu.cn}
\affiliation{
  \institution{Renmin University of China}
   \country{}
}
\authornotemark[1]
\author{Jingyu Liu}
\authornotemark[1]
\email{liujy1016@ruc.edu.cn}
\affiliation{%
  \institution{Renmin University of China}
  \country{}
  }
\author{Ge Chen}
\email{chenge221@mails.ucas.ac.cn}
\affiliation{%
  \institution{University of Chinese Academy of Sciences}
  \country{}
  }
\author{Zhirui Yang}
\email{yangzhirui@ruc.edu.cn}
\affiliation{%
  \institution{Renmin University of China}
  \country{}
  }

\author{Junchen Wan}
\email{wanjunchen@kuaishou.com}
\affiliation{%
  \institution{Kuaishou Technology}  
  \country{}
  }

\author{Fuzheng Zhang}
\email{zhangfuzheng@kuaishou.com}
\affiliation{%
  \institution{Kuaishou Technology}  
  \country{}
  }

\author{Zhongyuan Wang}
\email{wzhy@outlook.com}
\affiliation{%
  \institution{}  
  \country{}
  }

\author{Yong Liu}
\email{liuyonggsai@ruc.edu.cn}
\affiliation{%
  \institution{Renmin University of China}
  \country{}
  }

\begin{abstract}
Graph contrastive learning (GCL)  has emerged as a representative graph self-supervised method, achieving significant success. The currently prevalent optimization objective for GCL is InfoNCE. Typically, it employs augmentation techniques to obtain two views, where a node in one view acts as the anchor, the corresponding node in the other view serves as the positive sample, and all other nodes are regarded as negative samples. The goal is to minimize the distance between the anchor node and positive samples and maximize the distance to negative samples. However, due to the lack of label information during training, InfoNCE inevitably treats samples from the same class as negative samples, leading to the issue of false negative samples. This can impair the learned node representations and subsequently hinder performance in downstream tasks. While numerous methods have been proposed to mitigate the impact of false negatives, they still face various challenges. For instance, while increasing the number of negative samples can dilute the impact of false negatives, it concurrently increases computational burden. Thus, we propose GraphRank, a simple yet efficient graph contrastive learning method that addresses the problem of false negative samples by redefining the concept of negative samples to a certain extent, thereby avoiding the issue of false negative samples. The effectiveness of GraphRank is empirically validated through experiments on the node, edge, and graph level tasks. 
% The source code is provided here \url{https://anonymous.4open.science/r/GraphRank-18D1}.
\end{abstract}

 
\maketitle

\section{Introduction}~\label{intro}
Graph Neural Networks (GNNs) have become the standard approach for handling graph data, given their ability to leverage the underlying structure and features of graphs for effective analysis. Albeit the immense success achieved by supervised or semi-supervised GNNs~\cite{GCN,GAT} across numerous application domains, their effectiveness is tied to the availability of labeled data for learning robust and impactful node representations. However, obtaining labeled data in real-world scenarios is a costly and time-consuming endeavor, often constraining the availability of such data in many applications. Consequently, to mitigate this reliance on label data, graph self-supervised learning is attracting increasing attention, with graph contrastive learning emerging as the predominant method.
\begin{figure}
        \centering
        \includegraphics[width=0.8\linewidth]{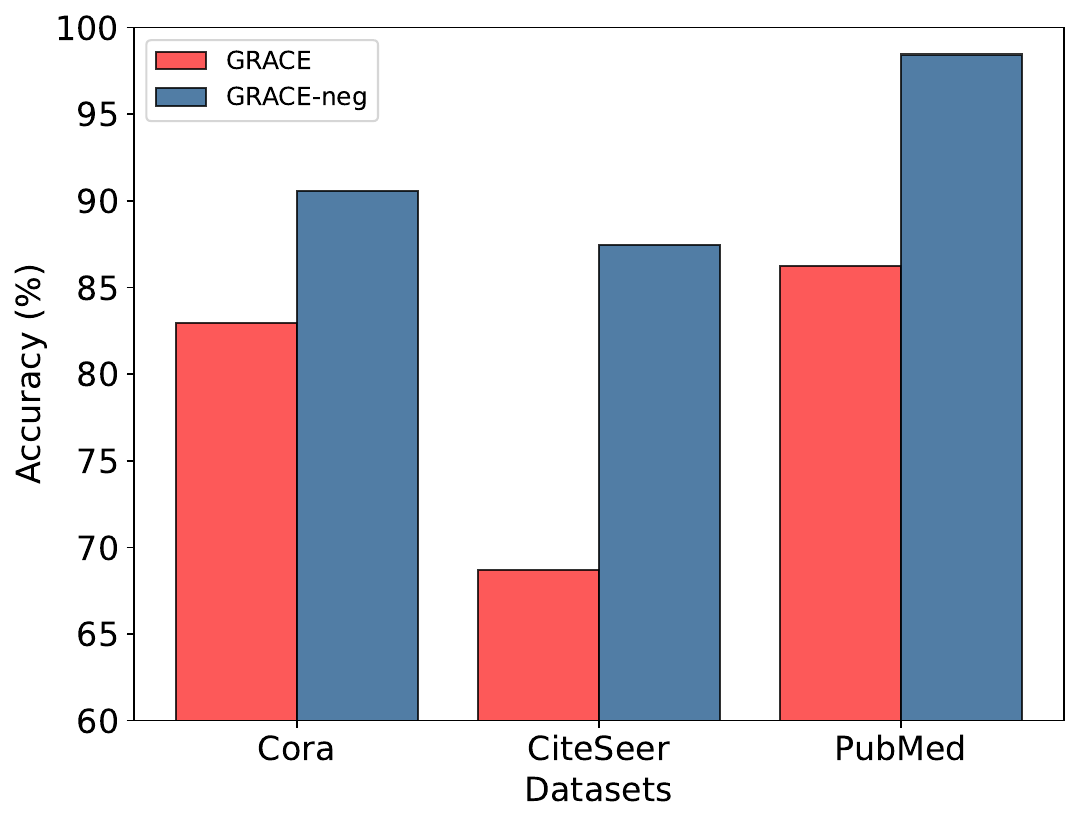}
        \caption{The impact of false negative samples. The legend "GRACE" denotes the results obtained by GRACE, while "GRACE-neg" represents the results derived from the GRACE with false negative samples excluded.}
        \label{figure_impact_of_intra_class_negative_nodes}
\end{figure}
 
% gcl->infonce-related work
Graph Contrastive Learning (GCL) typically starts with generating several views of a given graph through augmentation techniques. From these different views, one view serves as the anchor, with corresponding nodes in other views as positive examples and all other nodes as negative samples. The goal of GCL is then to bring the positive samples closer to the anchor in the representation space while pushing the negative samples further apart.
Among various GCL approaches, InfoNCE~\cite{qiu2020gcc,GRACE,GCL_GCA} has been recognized as the most commonly used optimization goal. It upholds the principle of minimizing distances between positive pairs and maximizing those between negative pairs, leveraging the contrastive nature of learning based on their representations. GRACE~\cite{GRACE} relies on hybrid feature augmentations including node feature masking and edge dropping. Based on this data augmentation strategy,  GCA~\cite{GCL_GCA} further introduces an adaptive augmentation for graph-structured data and make a competitive performance. GraphCL~\cite{GCL_GraphCL} further extends to graph-level representation to pull two views closer.

%weakness of infonce and existing solutions
However, InfoNCE experiences the issue of false negative samples~\cite{chuang2020debiased,niu2024affinity}. Its optimization objective is to minimize the distance with the positive samples and maximize the distance with the negative samples. Given that InfoNCE treats all nodes except the anchor node as negative samples, it unavoidably treats nodes of the same class as the anchor node as negative samples, which are referred to as false negative samples. Such false negative samples can impair the learned node representation and hinder downstream tasks. As shown in Figure~\ref{figure_impact_of_intra_class_negative_nodes}, GRACE, a representative graph contrastive learning method using InfoNCE loss, is used to validate the issue of false negative samples across three academic citation network datasets. %By deliberately removing false negative samples during the training of the GRACE method, we observed substantial performance improvements on all three datasets.
During the training of the GRACE method, we artificially removed the false negatives, which led to substantial performance improvements across all three datasets.
However, in practical scenarios, there is a lack of data label information at the time of training, preventing manual removal of false negative samples based on label information. 
% As such, numerous studies have explored ways to mitigate the issue of false negative samples and enhance the performance of graph contrastive learning methods, primarily in three areas.
Therefore, in order to alleviate the issue of false negative samples and improve the performance of graph contrastive learning methods, many works have been explored, which are mainly in three ways.
Firstly, increasing the number of negative samples helps dilute the impact of false negative samples. In cases where the quantities of various types of nodes are relatively balanced, the proportion of nodes in the same class is less, and increasing the number of negative samples can alleviate the issue of false negative samples.  However, an increase in the number of negative samples bears computational and storage burdens, and as shown in Figure~\ref{figure_impact_of_intra_class_negative_nodes}, GRACE, even when using all available negative samples, still experiences a notable false negative sample issue.
Secondly, some works have proposed mechanisms for screening negative samples to attempt to remove false negative samples, thus improving the quality of negative samples. AUGCL~\cite{chi2022enhancing} establishes a discriminative model based on collective affinity information to assess the uncertainty of negative samples, thereby facilitating the filtration of negative samples. Additionally, \cite{niu2024affinity} designes a mechanism based on node similarity to sample high-quality positive and negative samples. Although a negative sample screening mechanism can effectively reduce the sampling of false negative samples, it necessitates the design of a complex and intricate screening mechanism to ensure the selection of high-quality negative samples. This, in turn, would introduce additional computational overhead.
Lastly, some works have decided to forego the use of negative samples by employing contrastive learning methods that do not use negative samples, thereby avoiding the issue of false negative samples altogether. BGRL~\cite{thakoor2022largescale} is a contrastive learning method that does not require negative samples. It obtains two views through augmentation techniques; one view is used for learning the online representation, and the other view is used for learning the target representation. Updates are conducted by maximizing the similarity between these two views. %However, to stabilize training, BGRL requires a dual-encoder scheme with momentum update and exponential moving average features.
However, its success relies on a relatively complex training strategy, specifically requiring a dual-encoder scheme with momentum update and exponential moving average to stabilize the training process.

 %our work    
    In light of the shortcomings of these graph contrastive methods above, we propose a new framework for graph self-supervised learning called GraphRank. The GraphRank framework involves generating two augmented graph views by applying random masks to nodes and edges. Subsequently,we utilize a GNN as the encoder and employ rank loss as the objective function for training. Specifically, we select a node $v_i$ as the target node in view 1, the node $v_i^+$ corresponding to it in view 2 as a positive sample, and then randomly pick a node $v_j$ from view 2 as a negative sample. The representations of these nodes are derived by the encoder, and then the similarity between the target node and the positive and negative samples are calculated accordingly. By employing rank loss as the objective function, our aim is to ensure that the similarity between the target node and the positive samples is greater than the similarity between the target node and the negative samples.

    GraphRank can effectively address the problems mentioned above. Firstly, a simple random mask approach is applied to GraphRank to obtain augmented graph data, which does not require sophisticatedly designed graph augmentation techniques to obtain high-quality augmented graph data, nor does it require a complicated training strategy to stabilize the training. Secondly, we use rank loss as the objective function. Similar to the contrastive loss, e.g. InfoNCE, rank loss also endeavors to maximize the agreement between the target node and the positive sample. Different from contrastive loss, the purpose of rank loss is to make the similarity between the target node and the positive samples greater than the similarity between the target node and the negative samples, rather than separating the target node from the negative samples as much as possible, as in InfoNCE. Therefore, the rank loss would not separate the negative samples as far apart as possible, even if the negative samples selected were false negative samples. Finally, the calculation of rank loss involves only one positive and one negative sample resulting in a smaller computational overload compared to contrastive losses like InfoNCE. As a result, rank loss exhibits better scalability, making it more feasible for large-scale applications compared to typical contrastive losses.  
\section{Related Work}
 
\vpara{Contrastive-based SSL On Graph.}
GCL mainly contains three key components: data augmentation, GNN encoder and the contrastive loss. Most of recent works focus on the data augmentation process. DGI \cite{DGI}, GRACE \cite{GRACE}, GCC \cite{qiu2020gcc} generate corrupted graphs by corrupting input graphs with feature shuffle, removing the edges and masking the node features, and graph sampling respectively, while GCA \cite{GCL_GCA} calculates the importance of nodes/edges to conduct adative augmentation. And GraphCL \cite{GCL_GraphCL} selects different data augmentation strategies according to the domain of the dataset. Most works do no modification on the GNN encoder, some propose new contrastive loss \cite{CL_loss_corr,CL_loss_spec}, but most of them requires negative samples, and they all aim to align positive samples while separating negative samples \cite{GRACE,CL_simCLR}. Also they inevitably falls into the false negative problem, which means that there exists intra-class nodes in negative samples and the model mistakenly separates nodes of the same class. BGRL \cite{BGRL} and CCA-SSG \cite{CCA_SSG} both target a negative-sample-free model, but they still inevitably separates intra-class nodes, \cite{graph_neighbor_rank} remove the connected nodes from negative sets to relieve false negative problem, but the other intra-class negative samples are ignored. And most of those methods require high time complexity and fail in large graphs.

\vpara{Rank-based Approach on Graph.}
Learning to rank has been widely applied in various domains~\cite{li2011short,joachims2017unbiased,ji2014information}, such as information retrieval and natural language processing. However, its potential in the field of graph representation learning~\cite{ning2022graph,guo2023linkless} remains largely underexplored. In recent literature, GSCL~\cite{ning2022graph}  employs ranking techniques to filter high-quality positive samples. Specifically, they posit that the neighbors of an anchor node should maintain a high degree of similarity with the anchor node. As such, they consider neighbors as positive samples and utilize a ranking method to determine the importance of positive pairings.  Nevertheless, their method still relies on the InfoNCE loss for optimization. Another relevant study is LLP~\cite{guo2023linkless}, a graph knowledge distillation method proposed for link prediction tasks. They introduce a novel rank-based matching approach to distill relational knowledge from an teacher GNN into a student MLP. These methods merely utilize the ranking technique as an intermediate mechanism. In contrast, in this paper, we employ the ranking method to replace InfoNCE as the optimization objective for training.

\section{Limitations of Previous Researches}

\subsection{Preliminaries} We begin with some preliminary concepts and notations for further explanation. In this paper, $\mathcal{G}=(V,E)$ is used to represent a graph, and $V,E$ stands for its node set and edge set, respectively. We use $v_i$ to indicate the $i^{th}$ node and $X_{v_i} \in \mathbb{R}^F$ means its node attributes of dimension $F$. $A \in \mathbb{R}^{n \times n}$ is the adjacency matrix, while $n=|V|$ stands for the number of nodes, $A_{i,j}=1$ iff $(v_i,v_j)\in E$ \ie node $v_i$ and $v_j$ are connected.

    \vpara{Graph Contrastive Learning.} Graph Contrastive Learning (GCL) commonly involves generating two augmented views, denoted as $\mathcal{G}^1$ and $\mathcal{G}^2$, and try to maximize the mutual information or the correspondence between two different views \cite{GCL_Information}:
    \begin{equation}\label{equation:MI}
        \text{InfoMax}: \mathrm{max}_{f \in \mathcal{F}} \  \mathrm{MI}(f(\mathcal{G}^1),f(\mathcal{G}^2)),
    \end{equation}
    where MI stands for the mutual information, $f$ is the graph encoder, and $\mathcal{F}$ is the function class.
    
    In order to optimize Equation (\ref{equation:MI}), GCL methods regard nodes augmented from the same as positive pair, and others as negative pair, and GCL is trained to maximize the similarity between positive pairs and minimize the similarity between negative ones. For instance, the most widely used loss function InfoNCE loss is defined as below \cite{CL_simCLR}:

    \begin{equation}\label{infoNCE}
        \mathcal{L}_{\mathrm{NCE}}=\E_{p(v_i,v_i^+)}\E_{\{ p(v_i^-) \}} \left[-\log \frac{\exp(f(v_i)^Tf(v_i^+) / \tau)}{\sum_{i=1}^M\exp(f(v_i)^Tf(v_i^-) / \tau)}\right],
        \nonumber
    \end{equation}
    where $v_i^+$ stands for the node augmented from $v_i$ \ie positive pair, $v^-$ stands for randomly sampled nodes \ie negative pair and $\tau$ is the temperature.
    
    InfoNCE loss is proved to be a lower bound of mutual information \cite{NCE_MI}, minimizing InfoNCE effectively maximizes the mutual information between two views. This property has contributed to the success of various contrastive algorithms in achieving satisfactory results. However, InfoNCE loss tries to minimize the similarity between negative pairs even when the negative pair belongs to the same class which will inevitably reduce the downstream performance \cite{CL_imperfect_alignment}. Also previous theoretical researches \cite{CL_negative_sample_number} and algorithms \cite{GRACE} all indicate that incorporating more negative samples can improve performance in contrastive learning tasks, but it will also leads to larger computational complexity, making it challenging to deploy such methods on large graphs.
    
\subsection{Intra-class Negative Nodes}
    % \begin{figure}
    %     \centering
    %     \includegraphics[width=0.8\linewidth]{images/erase_intra_negative_GRACE.pdf}
    %     \caption{The impact of intra-class negative nodes}
    %     \label{figure_impact_of_intra_class_negative_nodes}
    % \end{figure}
    Most previous researches focus on how to perform augmentation and how to sample the hard negative nodes \cite{GCL_hard_negative}, the intra-class negative problem \ie false negative \cite{CL_first} is largely overlooked. In fact contrastive learning methods all inevitably sample intra-class nodes as negative samples because the class label information is not accessible during the pretraining, as a result, the model mistakenly separates intra-class nodes away, leading to suboptimal results.
    
    Figure \ref{figure_impact_of_intra_class_negative_nodes} demonstrates the impact of removing intra-class negative nodes on the downstream performance of the model. By manually removing all intra-class negative nodes, the downstream performance would be much better as the model only decreases the similarity of inter-class nodes, and intra-class nodes will gather closer as the model would not separate any intra-class nodes anymore.
    Therefore, the existence of intra-class negative nodes significantly diminish downstream performance, and there are some works noticing this problem \cite{CL_weak_supervised,CL_recong_false_neg}, but they all try to solve the problem by probing those intra-class negatives and remove them from the negative sample set. Different from previous works, we think it is the widely used InfoNCE loss deepening the impact of intra-class negatives as it requires to push intra-class negative pairs further and further.

    \begin{figure*}
        \centering
        \includegraphics[width=0.8\linewidth]{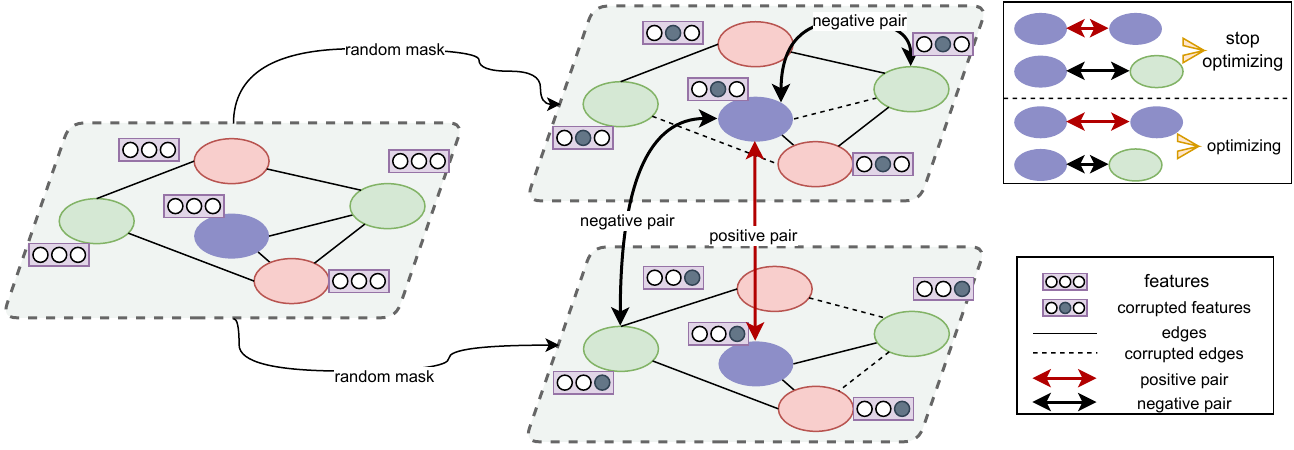}
        \caption{Model architecture}
        \label{figure_model_architecture}
    \end{figure*}
 
\subsection{Computational Complexity} \label{sec:complexity}

    As algorithms like GRACE \cite{GRACE} requires to separate negative nodes as far as possible \cite{CL_simCLR}, if we only sample few negative nodes, for example, for an anchor node $v_1$ only 1 negative node $v_2$ is sampled, then the model with InfoNCE loss would try to minimize the similarity between $v_1$ and $v_2$. However, if the negative pair $v_1$, $v_2$ are of the same class, the model may inadvertently push intra-class nodes apart, leading to poor downstream performance.

    Therefore, contrastive learning algorithms always require a substantial number of negative samples \cite{CL_negative_sample_number,GRACE}. Because large number of inter-class negative samples could alleviate the impact of intra-class negative samples. However, having more negative samples also translates to increased computational complexity. For instance, GRACE uses all other nodes as negative samples, and resulting in a time complexity of $O(N^2)$. So GRACE can hardly be deployed on large graphs, if we reduce the number of negative samples to mitigate this issue, it would lead to a sharp decrease in performance.

\begin{table}[tb!]
	\centering
	\caption{Technical comparison of self-supervised node representation learning methods. \textit{Target} denotes the comparison pair, N/G/F denotes node/graph/feature respectively. \textit{False neg} means does the method face the false negative problem. \textit{More negs} stands for how much negative samples are needed. \textit{Time} and \textit{Space} is the time and space complexity respectively.}
	\label{tbl-comparison}
	\small
	\begin{threeparttable}
    {{
    		\begin{tabular}{c|lcccccc}
			\toprule[0.8pt]
			& Methods & Target & \begin{tabular}[c]{@{}c@{}}False \\ neg\end{tabular} & \begin{tabular}[c]{@{}c@{}}More \\ negs\end{tabular} & Time & Space  \\
			\midrule[0.6pt]
           \multirow{6}{*}{\rotatebox{90}{Instance-level}} 
            & DGI     & N-G  &- &- &$O(N)$ &$O(N)$\\
            & GRACE   & N-N &\checkmark & \checkmark &$O(N^2)$ &$O(N^2)$ \\
            & GCA   & N-N &\checkmark & \checkmark &$O(N^2)$ &$O(N^2)$ \\
            & GraphCL      & N-N &\checkmark & \checkmark &$O(N^2)$ &$O(N^2)$ \\
            &GCC &G-G &- &\checkmark &$O(N^2)$ &$O(N^2)$ \\
            &CCA-SSG &F-F &- &\checkmark &$O(N)$ &$O(N)$ \\
            \midrule[0.5pt]
            & GraphRank & N-N  &- &- &$O(N)$ &$O(N)$ \\
			\bottomrule[0.8pt]
		\end{tabular}
		}
		}
	\end{threeparttable}
\end{table}

\section{Methodology}

In this section, we will provide a detailed introduction to  GraphRank framework. As depicted in Figure~\ref{figure_model_architecture}, GraphRank operates by generating two views of the original graph through random masking. Subsequently, we train the model using the rank loss objective, which aims to enhance the similarity between the representations of corresponding nodes from the two views, while ensuring a greater dissimilarity with the representations of other nodes.

\subsection{Rank Loss}
    As we mentioned before, the traditional contrastive loss all encounters the challenge of handling intra-class negative samples. When a negative sample is of the same class, conventional loss still try to minimize the similarity, which would definitely reduce the downstream performance. Therefore, what we require is an optimization objective that promotes similarity between positive pairs while avoiding excessive separation of negative pairs. The rank loss precisely fulfills these requirements and aligns with our objectives.
    \begin{equation}\label{equation:rank_loss}
        \mathcal{L}_{\mathrm{rank}}=\mathrm{max} \{ 0, \mathrm{margin}-(\mathrm{sim}(v,v^+)-\mathrm{sim}(v,v^-)) \},
    \end{equation}
    where $\mathrm{sim}$ stands for a function that evaluate the similarity between two inputs, $\mathrm{margin}$ is a hyperparameter.

    For Equation (\ref{equation:rank_loss}), if we simply set $\mathrm{margin}$ to $0$, then we can observe that, when the similarity of a positive pair is higher than that of a negative pair, the loss on this node would be 0, and when negative pair gets higher similarity, the loss would be $\mathrm{sim}(v,v^-)-\mathrm{sim}(v,v^+)$, and the model would try to minimize the value \ie increasing $\mathrm{sim}(v,v^+)$ and decreasing $\mathrm{sim}(v,v^-)$ until $\mathrm{sim}(v,v^+) \geq \mathrm{sim}(v,v^-)$.
    
    Indeed, the hyperparameter $\mathrm{margin}$ plays a crucial role in the rank loss. It determines the degree to which the similarity of positive pairs should exceed that of negative pairs. By setting an appropriate value for $\mathrm{margin}$, the rank loss can effectively separate negative nodes. When the learned representations for a negative pair are already sufficiently distant, the loss is set to 0, allowing the training process to focus on more challenging pairs and further enhancing their separation. This ensures that the model allocates its efforts towards optimizing pairs that require additional optimization, leading to better performance on those specific nodes. Selecting an appropriate value for $\mathrm{margin}$ is essential for achieving the desired balance between encouraging positive pairs and separating negative pairs in the rank loss.

    Unlike the InfoNCE loss, the Rank loss does not enforce the similarity of negative pairs to be minimized as much as possible. Instead, it only requires that the similarity of the positive pairs is a little higher than the similarity of the negative pairs. It ensures the separation between different classes while prevents intra-class nodes being separated too far. By adopting this approach, the Rank loss allows positive pairs to be brought closer together, effectively mitigating the negative impact of intra-class negative samples.
    
\subsection{SSL with Rank Loss.} In Figure \ref{figure_model_architecture}, we illustrate our unsupervised graph contrastive learning framework, called GraphRank. Similar to typical contrastive learning methods, we generate two augmented views by randomly corrupting the original graph, and we also regard the nodes augmented from the same as positive pair \ie $v^1,v^2$ and others as negative pair \ie $v^1,v^-$. Then we use the rank loss to assure: 
    \begin{equation}
        \mathrm{sim}(f(v^1),f(v^2)) - \mathrm{sim}(f(v^1),f(v^-)) \geq \mathrm{margin}.
    \end{equation}
    % $$\mathrm{sim}(f(v^1),f(v^2)) - \mathrm{sim}(f(v^1),f(v^-)) \geq \mathrm{margin}$$.

    Similar to GRACE \cite{GRACE}, we generate augmented views by removing edges and masking node features. Specifically, we randomly remove a portion of edges in the original graph by formulating a masking matrix $\bm{M}_e$, whose entry is drawn from a Bernouli distribution $\widetilde{\bm{R}}_{ij} \sim \mathcal{B}(1-p_{e})$ if $\bm{A}_{ij}=1$, and $\widetilde{\bm{R}}_{ij}=0$, otherwise. Also we randomly mask some node features by sampling a random vector $\widetilde{\bm{m}} \in \{0,1\}^F$ where each dimension of vector $\widetilde{\bm{m}}$ is drawn from a Bernouli distribution $\widetilde{\bm{m}} \sim \mathcal{B}(1-p_{f})$. And we set the masking probability $p_e$ and $p_f$ differently when augmenting two views.

    After the augmentation process, we employ a Graph Neural Network (GNN) to learn the embedding of nodes. For each node $v_i$, we locate its positive pair $v_i^+$, and randomly sample a node $v_i^-$ as the negative pair, we calculate the positive/negative pair similarity by $f(v_i)^Tf(v_i^+)$ and $f(v_i)^Tf(v_i^-)$. Next, we use the rank loss to align positive pair and separate negative ones. Like we mentioned before, by using rank loss, we do not require negative pair similarity to be as small as possible, and the model could focus more on those negative nodes who need further optimization. Moreover, the rank loss inherently accounts for intra-class negative nodes, as it only requires the similarity of positive pairs to be higher than that of negative pairs, which is axiomatic no matter the negative node is of the same class or not. 
    The learning algorithm is summarized as below:
    \begin{algorithm}
        \caption{GraphRank training algorithm} \label{algorithm_GraphRank}
        \KwData{the graph $\mathcal{G}$, graph encoder $f$}
        \KwResult{node embedding of the original graph $f(\mathcal{G})$}
        
        \For {$\mathrm{epoch} \leftarrow 1,2,... $}
        {
            Generate two graph views $\mathcal{G}^1$, $\mathcal{G}^2$ by random corruption on $\mathcal{G}$;
            
            Obtain node embeddings of both views $f(\mathcal{G}^1)$, $f(\mathcal{G}^2)$;
            
            Randomly sample 1 negative pair for each node in $\mathcal{G}^1$;
            
            Compute the rank loss with Equation \ref{equation:rank_loss};
            
            Update parameters using stochastic gradient ascent;
            
        }
        Obtain node embeddings of the original view $f(\mathcal{G})$ by the final parameter.
    \end{algorithm}

\subsection{Rank Loss Benefits SSL}

\begin{figure*}[h]
    \centering
    \begin{subfigure}[b]{0.31\textwidth}
        \includegraphics[width=\textwidth]{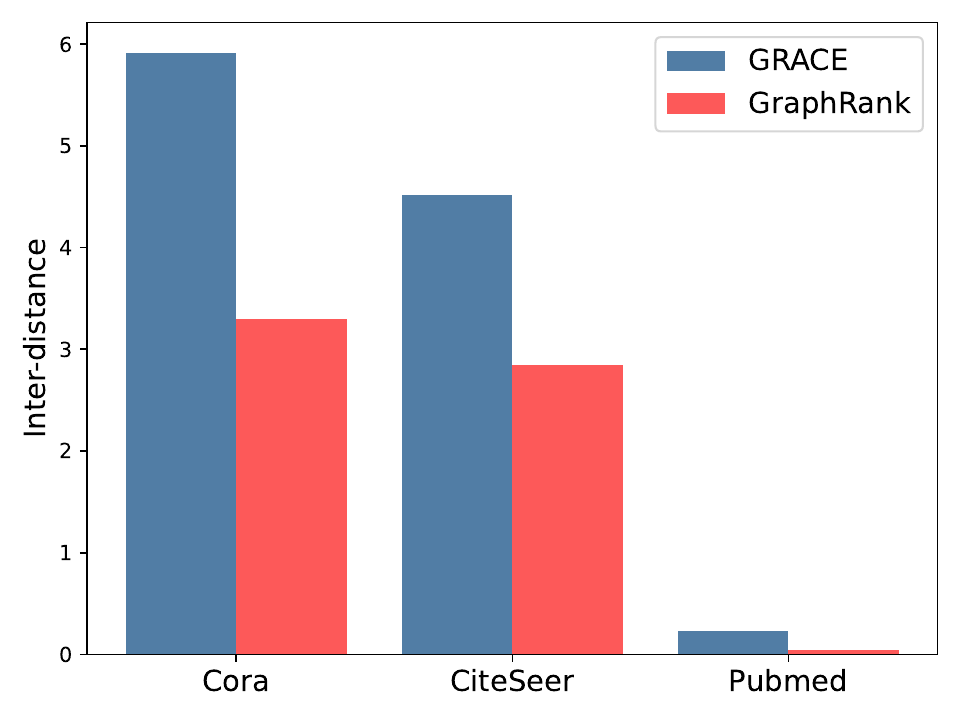}
        \caption{Intra-class variance}
        \label{figure_intra-class_variance}
    \end{subfigure}
    \hfill 
    \begin{subfigure}[b]{0.335\textwidth}
        \includegraphics[width=\textwidth]{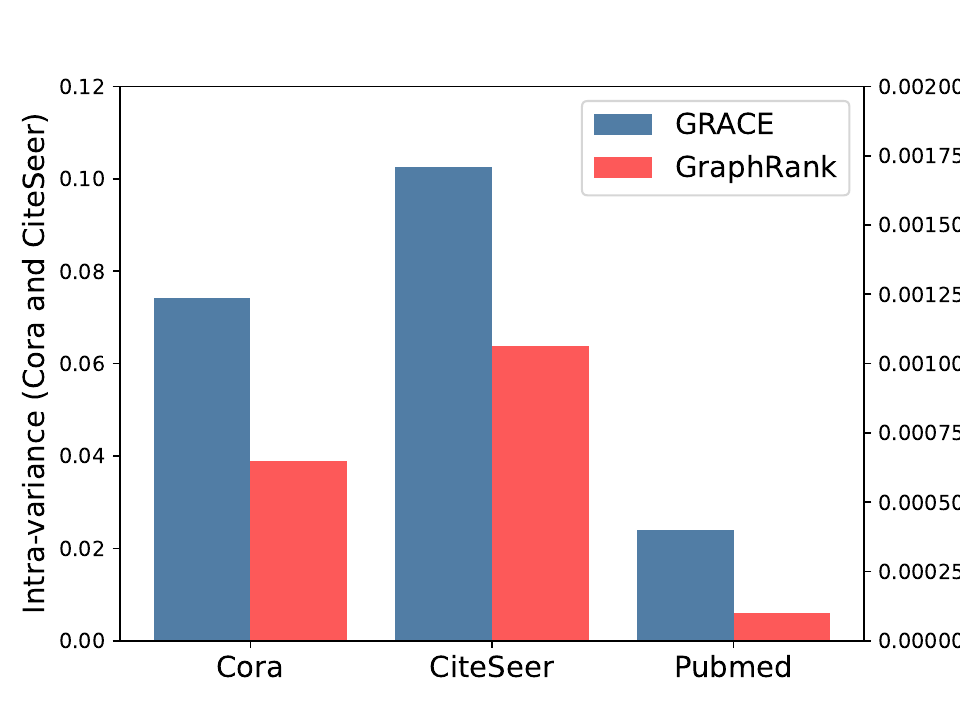}
        \caption{Inter-class distance}
        \label{figure_inter-class_distance}
    \end{subfigure}
    \hfill
    \begin{subfigure}[b]{0.33\textwidth}
        \includegraphics[width=\textwidth]{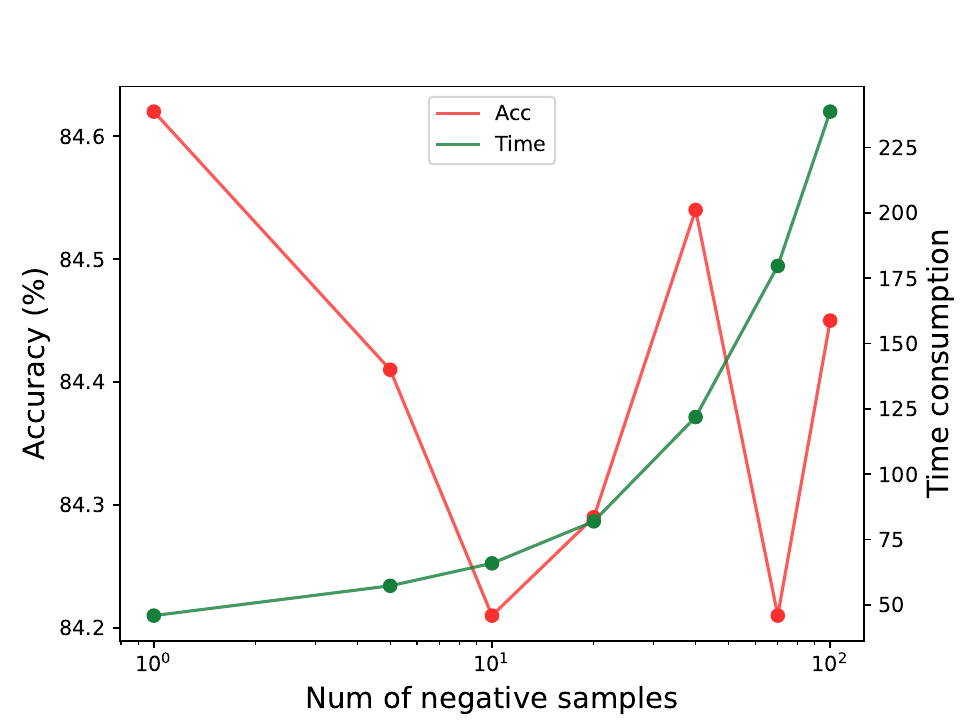}
        \caption{Impact of negative sample number}
        \label{figure_impact_of_negative_number_rank}
    \end{subfigure}
    \caption{Analysis of rank loss}
    \label{fig:tri_figure}
\end{figure*}

\vpara{Intra-class Variance.} In GraphRank, the focus is on preventing intra-class negative nodes from being excessively separated, which leads to positive nodes of the same class being brought closer together compared to the commonly used InfoNCE loss. As shown in Figure \ref{figure_intra-class_variance}, GraphRank achieves significantly smaller intra-class variance compared to GRACE \cite{GRACE}, which means GraphRank gathers nodes of the same class closer, and results in a better downstream performance.

    Also, GraphRank no longer pushes intra-class nodes away does not only means a smaller intra-class variance, more importantly, it greatly reduces the number of outliers which directly helps the downstream tasks as the outliers could be classified/predicted more precisely.

\vpara{Inter-class Distance.} \label{section:inter-class_distance}
In the absence of class label information during pretraining, GraphRank cannot exclusively prevent the separation of intra-class negative samples. Inter-class negative samples would also inevitably be less separated, which would result in a lower inter-class distance. However, the hyperparameter $\mathrm{margin}$ could control how low we want negative pair similarity to be, so inter-class nodes could still be pushed away to some distance, but still GraphRank can hardly separates different classes as far as the InfoNCE loss.   
    However, experiments shown in Section \ref{section_experiments} indicates that, suboptimal performance on separating inter-class nodes has a limited impact on downstream tasks. GraphRank may learn a shorter inter-class distance, but it still outperforms other GCL methods based on InfoNCE who gets longer inter-class distance. This is because although GraphRank can not separates different classes very far, it still keeps a clear boundary between different classes while intra-class nodes gather much closer including those outliers, the outstanding performance on node clustering shown in Table~\ref{tab:cluster} also prove this. Therefore, GraphRank may lead to lower inter-class node distance, but it is still clear to recognize different classes and the gathering of intra-class nodes especially the outliers benefits SSL. Further elaboration on the definitions and calculations of Intra-class Variance and Inter-class Distance is provided in Appendix~\ref{app:remark}.

\vpara{Few Negative Sample Learning.}
    Previous works that utilize contrastive loss, such as InfoNCE, often require a large number of negative samples. This is mainly because, with a small number of negative samples, such as only one, if that sample happens to be from the same class as the anchor node, the InfoNCE loss would minimize the similarity, leading to an extremely large intra-class distance. To mitigate this effect, more negative samples are needed to dilute the impact of intra-class negatives. However, GraphRank set a clip for the difference between positive and negative pair similarity, so it will not draw the intra-class negative too far as long as we set a reasonable margin. Consequently, GraphRank is less affected by accidentally occurring intra-class negative nodes, even when the number of negative samples is small. Like we show in Figure \ref{figure_impact_of_negative_number_rank}, GraphRank could achieve satisfying downstream performance with only few negative samples. This reduction in the number of required negative samples greatly reduces the computational complexity of GraphRank, making it more efficient and practical for real-world applications.

    To further evaluate the performance of GraphRank with few negative samples, we sample only 1 negative sample in Section \ref{section_experiments} if not specifically pointed out.

\section{Experiments}\label{section_experiments}
% \begin{table}[!htbp]
%       \caption{Dataset statistics}
%       \label{Dataset_statistics}
%       \centering
%       \setlength{\tabcolsep}{1mm}{
%         \begin{tabular}{ccccc}
%             \toprule 
%             Dataset & Nodes & Edges & Features & Classes \\
%             \midrule
%             Cora  & 2,708 & 5,429 & 1,433 & 7 \\
%             Citeseer  & 3,327 & 4,732 & 3,703 & 6\\
%             Pubmed  & 19,717 & 44,338 & 500 & 3\\
%             Photo  & 7,650 & 119,081 & 745 & 8 \\
%             Computers  & 13,752 & 245,861 & 767 & 10 \\
%             CS &18,333 &182,121 &6,805 &15\\
%             Physics &34,493 &530,417 &8,415 &5\\
%             \bottomrule
%         \end{tabular}
%     }
%     \end{table}

\begin{table*}[tb]
\caption{Experimental results of node classification. The best results are marked bold,  while the second-best results are underlined.}
\label{tab:nc}
\begin{threeparttable}
    
\begin{tabular}{llllllll}
\hline
Methods   & Cora                                         & Citeseer                                     & Pubmed                                       & Photo                                                       & Computer                                                       & CS                                                    & Physics                                                     \\ \hline
GCN       & 80.84±0.44                                   & 70.46±0.85                                   & 79.02±0.26                                   & 90.79±2.47                                                  & 85.34±1.64                                                     & 92.10±0.16                                            & 95.41±0.11                                                  \\
GAT       & 83.00±0.70                                   & 72.50±0.70                                   & 79.00±0.30                                   & 93.49±0.16                                                  & 89.79±0.76                                                     & 91.48±0.17                                            & 95.25±0.14                                                  \\
APPNP     & 83.70±0.25                                   & 72.10±0.30                                   & 79.73±0.31                                   & 93.42±0.35                                                  & 87.45±0.66                                                     & 92.50±0.14                                            & 95.59±0.09                                                  \\ \hline
GRACE     & 77.60±0.28                                   & 67.24±0.93                                   & 78.40±1.33                                   & 92.45±0.34                                                  & 88.11±0.45                                                     & 67.57±12.91                                           & 85.33±6.76                                                  \\
DGI       & 82.30±0.60                                   & 71.80±0.70                                   & 76.80±0.60                                   & 91.61±0.22                                                  & 83.95±0.47                                                     & 92.15±0.63                                            & 94.51±0.52                                                  \\
CCA-SSG   & 84.00±0.40                                   & 73.10±0.30                                   & 81.00±0.40                                   & 93.14±0.14                                                  & 88.74±0.28                                                     & 93.31±0.22                                            & 95.38±0.06                                                  \\
MVGRL     & 83.50±0.40                                   & 73.30±0.50                                   & 80.10±0.70                                   & 91.74±0.07                                                  & 87.52±0.11                                                     & 92.11±0.12                                            & 95.33±0.03                                                  \\
GraphMAE  & 84.20±0.40                                   & {\ul 73.40±0.40}                             & {\ul 81.10±0.40}                             & 92.98±0.35                                                  & 88.34±0.27                                                     & 93.08±0.17                                            & 95.30±0.12                                                  \\
BGRL      & \multicolumn{1}{c}{{\color[HTML]{1F2329} -}} & \multicolumn{1}{c}{{\color[HTML]{1F2329} -}} & \multicolumn{1}{c}{{\color[HTML]{1F2329} -}} & \multicolumn{1}{c}{{\color[HTML]{1F2329} {\ul 93.17±0.30}}} & \multicolumn{1}{c}{{\color[HTML]{1F2329} \textbf{90.34±0.19}}} & \multicolumn{1}{c}{{\color[HTML]{1F2329} 93.31±0.13}} & \multicolumn{1}{c}{{\color[HTML]{1F2329} {\ul 95.73±0.05}}} \\
SeeGera   & {\ul 84.30±0.40}                             & 73.00±0.80                                   & 80.40±0.40                                   & 92.81±0.45                                                  & 88.39±0.26                                                     & {\ul 93.84±0.11}                                      & 95.39±0.08                                                  \\ \hline
GraphRank & \textbf{84.93±1.07}                          & \textbf{73.43±0.93}                          & \textbf{85.75±0.16}                          & \textbf{93.60±0.18}                                         & {\ul 89.48±0.22}                                               & \textbf{94.76±0.08}                                   & \textbf{96.28±0.06}                                         \\ \hline
\end{tabular}
\begin{tablenotes}
      % \item {"-" denotes results that were not reported in the original paper..}
      \footnotesize
    \item[] \scriptsize "-" indicate tha the results not reported in the original paper.
    \end{tablenotes}
\end{threeparttable}
\end{table*}

    In this section, we conduct extensive experiments to evaluate the effectiveness and efficiency of GraphRank on both node classification and link prediction tasks. The experiments are designed to address the following research questions ($\mathcal{RQ}$):
    \begin{description}
        \item[$\mathcal{RQ}1$]: How does GraphRank compare to other self-supervised and semi-supervised methods on various graph tasks?
        \item[$\mathcal{RQ}2$]: How is the training efficiency of GraphRank and its sensitivity to hyperparameters?
    \end{description}

\subsection{Experimental Setup}
    \vpara{Datasets.} For node classification and link prediction tasks, the experiments are mainly conducted on 7 datasets: Cora, CiteSeer, PubMed \cite{dataset_Cora_CiteSeer_PubMed2}, Photo, Computer, CS and Physics \cite{dataset_Amazon_P_C}. The first three datasets \cite{dataset_Cora_CiteSeer_PubMed2} are citation networks, Photo and Computer \cite{dataset_Amazon_P_C} are derived from the Amazon product co-purchasing network, while Nodes in CS and Physics represent the author, the link stands for coauthor relationship between authors.  
    For the graph classification datasets, we conduct experiments on seven datasets: MUTAG, IMDB-B, IMDB-M, PROTEINS, COLLAB and REDDITB~\cite{yanardag2015deep}.
    More information about the datasets can be found in Appendix~\ref{data}.

    \vpara{Baselines.} For node classification and link prediction tasks, some supervised and unsupervised graph neural networks are used as baseline: {Supervised methods}, including GCN \cite{GCN}, GAT \cite{GAT} and APPNP\cite{APPNP}. {Generative graph SSL methods}, GraphMAE \cite{GraphMAE} CAN \cite{CAN}, SIG-VAE \cite{SIG_VAE},  and SeeGera \cite{SeeGera} are categorized as generative methods. {Contrastive methods}, including GRACE \cite{GRACE}, DGI \cite{DGI}, CCA-SSG \cite{CCA_SSG}, MVGRL \cite{MVGRL}. 
For graph classification tasks, the baselines we compare against can be divided into three main categories:(1){Supervised methods}, including GIN~\cite{xu2018powerful} and DiffPool. (2) Graph kernel methods, which comprise of WeisfeilerLehman sub-tree kernel (WL)~\cite{shervashidze2011weisfeiler} and Deep Graph Kernel (DGK)~\cite{yanardag2015deep}. (3) Self-supervised methods, which include graph2vec, Infograph~\cite{sun2019infograph}, GraphCL~\cite{you2020graph}, JOAO~\cite{you2021graph}, GCC~\cite{qiu2020gcc}, MVGRL, BGRL and InfoGCL~\cite{xu2021infogcl}
    
    \vpara{Settings and Hyperparameters.} For all baselines, we ensure consistency with the setup described in \citep{SeeGera}. We reproduce the experiments using the official code whenever available. In cases where official code is not provided, we report the results as presented in the original paper. In the case of GraphRank, we employ a two-layer GCN or GAT as the encoder, and the mask ratio of nodes and edges ranges from 0.1 to 0.5. The value of margin in Equation~\ref{equation:rank_loss} is generally fixed to 0. For the node classification task and graph classification task, we measure performance in terms of accuracy (Acc). For the link prediction task, we follow the settings of SEEGERA~\citep{SeeGera} where 10\% of the edges existed in the original graph are randomly selected as positive examples and an equal number of non-existing edges are chosen as negative examples to form the test set. We utilize area under the ROC curve (AUC) and the average precision (AP) as evaluation metrics to measure the performance of link prediction. All experiments are conducted on a single V100 32G. We repeated each experiment five times and report the mean and standard deviation of the results.
    %For the link prediction task, we use area under the ROC curve (AUC) and the average precision (AP) as evaluation metrics.

\subsection{Performance Evaluation}

\begin{table*}[!htb]
\caption{Experimental results of link prediction. The best results are marked bold,  while the second-best results are underlined.}
\label{tab:LP}
\begin{threeparttable}

\begin{tabular}{lllcccccc}
\hline
                                           & Method    & Cora                  & \multicolumn{1}{l}{Citeseer} & \multicolumn{1}{l}{Pubmed} & \multicolumn{1}{l}{Photo} & \multicolumn{1}{l}{Computer} & \multicolumn{1}{l}{CS}  & \multicolumn{1}{l}{Physics} \\ \hline
\multicolumn{1}{l|}{\multirow{10}{*}{AUC}} & DGI       & 93.88±1.00        & 95.98±0.72               & 96.30±0.20             & 80.95±0.39            & 81.27±0.51               & 93.81±0.20          & 93.51±0.22              \\
\multicolumn{1}{l|}{}                      & MVGRL     & 93.33±0.68        & 88.66±5.27               & 95.89±0.22             & 69.58±2.04            & 92.37±0.78               & 91.45±0.67          & OOM                         \\
\multicolumn{1}{l|}{}                      & GRACE     & 82.67±0.27        & 87.74±0.96               & 94.09±0.92             & 81.72±0.31            & 82.94±0.20               & 85.26±2.07          & 83.48±0.96              \\
\multicolumn{1}{l|}{}                      & GCA       & 81.46±4.86        & 84.81±1.25               & 94.20±0.59             & 70.02±9.66            & 89.92±0.91               & 84.35±1.13          & 85.24±5.41              \\
\multicolumn{1}{l|}{}                      & CCA-SSG   & 93.88±0.95        & 94.69±0.95               & 96.63±0.15             & 73.98±1.31            & 75.91±1.50               & 96.80±0.16          & 96.74±0.05              \\
\multicolumn{1}{l|}{}                      & CAN       & 93.67±0.62        & 94.56±0.68               & -                          & {\ul 97.00±0.28}      & {\ul 96.03±0.37}         & -                       & -                           \\
\multicolumn{1}{l|}{}                      & SIG-VAE   & 94.10±0.68        & 92.88±0.74               & 85.89±0.54             & 94.98±0.86            & 91.14±1.10               & 95.26±0.36          & {\ul 98.76±0.23}        \\
\multicolumn{1}{l|}{}                      & GraphMAE  & 90.70±0.01        & 70.55±0.05               & 69.12±0.01             & 77.42±0.02            & 75.14±0.02               & 91.47±0.01          & 87.61±0.02              \\
\multicolumn{1}{l|}{}                      & SEEGERA   & {\ul 95.50±0.71}  & {\ul 97.04±0.47}         & {\ul 97.87±0.20}       & \textbf{98.64±0.05}   & \textbf{97.70±0.19}      & \textbf{98.42±0.13} & \textbf{99.03±0.05}     \\ \cline{2-9} 
\multicolumn{1}{l|}{}                      & GraphRank & \textbf{97.74±0.41}   & \textbf{98.49±0.15}          & \textbf{98.77±0.07}        & 95.31±0.33                & 95.59±0.87                   & {\ul 98.35±0.37}        & 96.75±0.07                  \\ \hline
\multicolumn{1}{l|}{\multirow{10}{*}{AP}}  & DGI       & 93.60±1.14        & 96.18±0.68               & 95.65±0.26             & 81.01±0.47            & 82.05±0.50               & 92.79±0.31          & 92.10±0.29              \\
\multicolumn{1}{l|}{}                      & MVGRL     & 92.95±0.82        & 89.37±4.55               & 95.53±0.30             & 63.43±2.02            & 91.73±0.40               & 89.14±0.93          & OOM                         \\
\multicolumn{1}{l|}{}                      & GRACE     & 82.36±0.24        & 86.92±1.11               & 93.26±1.20             & 81.18±0.37            & 83.12±0.23               & 83.90±2.20          & 82.20±1.06              \\
\multicolumn{1}{l|}{}                      & GCA       & 80.87±4.11        & 81.93±1.76               & 93.31±0.75             & 65.17±10.11           & 89.50±0.64               & 83.24±1.16          & 82.80±4.46              \\
\multicolumn{1}{l|}{}                      & CCA-SSG   & 93.74±1.15        & 95.06±0.91               & 95.97±0.23             & 67.99±1.60            & 69.47±1.94               & 96.40±0.30          & 96.26±0.10              \\
\multicolumn{1}{l|}{}                      & CAN       & 94.49±0.60        & 95.49±0.61               & -                          & {\ul 96.68±0.30}      & {\ul 95.96±0.38}         & -                       & -                           \\
\multicolumn{1}{l|}{}                      & SIG-VAE   & 94.79±0.71        & 94.21±0.53               & 85.02±0.49             & 94.53±0.93            & 91.23±1.04               & 94.93±0.37          & {\ul 98.85±0.12}        \\
\multicolumn{1}{l|}{}                      & GraphMAE  & 89.52±0.01        & 74.50±0.04               & 87.92±0.01             & 77.18±0.02            & 75.80±0.01               & 83.58±0.01          & 86.44±0.03              \\
\multicolumn{1}{l|}{}                      & SEEGERA   & {\ul 95.92±0.68}  & {\ul 97.33±0.46}         & {\ul 97.87±0.20}       & \textbf{98.48±0.06}   & \textbf{97.50±0.15}      & \textbf{98.53±0.18} & \textbf{99.18±0.04}     \\ \cline{2-9} 
\multicolumn{1}{l|}{}                      & GraphRank & \textbf{97.98±0.33} & \textbf{98.28±0.28}        & \textbf{98.39±0.11}        & 94.65±0.35              & 95.08±0.69                 & {\ul 98.14±0.38}      & 96.22±0.05                \\ \hline
\end{tabular}
\begin{tablenotes}
    \footnotesize
    \item[] \scriptsize The symbol "-" denotes unreported results due to either unavailability of the code or out-of-memory. 
      % \item {The symbol "-" denotes unreported results due to either unavailability of the code or out-of-memory.}
    \end{tablenotes}
\end{threeparttable}
\end{table*}

\vpara{Node Classification.} We conducted a comprehensive comparison of GraphRank with three supervised methods and six self-supervised methods across seven datasets. The experimental results for node classification are presented in Table~\ref{tab:nc}. Self-supervised methods can outperform supervised methods or achieve comparable performance to them, which highlights the effectiveness and great potential of graph self-supervision methods for learning graph data representations. GraphRank stands out by outperforming most of the compared supervised and self-supervised baselines on six out of the seven datasets. On the Computer dataset, GraphRank also achieves comparable result against the best baseline. This consistent and impressive performance across diverse datasets validates the effectiveness and generalizability of GraphRank. Furthermore, GraphRank demonstrates superior performance compared to contrastive learning methods like GRACE. This outcome supports our analysis that GraphRank effectively mitigates the impact of intra-class negative nodes, leading to the learning of enhanced node representations that greatly benefit downstream tasks.

\vpara{Link Prediction.}
We compare GraphRank with 9 SSL baselines, including the contrastive methods: DGI~\cite{DGI}, MVGRL~\cite{MVGRL}, GRACE~\cite{GRACE}, GCA~\cite{GCL_GCA}, CCA-SSG~\cite{CCA_SSG}, and the generative methods: CAN~\cite{CAN}, SIG-VAE~\cite{SIG_VAE}, GraphMAE~\cite{GraphMAE}, SeeGera~\cite{SeeGera}. In the link prediction task, we follow the approach of SeeGera to calculate the predicted probability of an edge. Specifically, for an edge between nodes $v_i$ and $v_j$, the predicted probability $p_{ij}$ is computed as $p_{ij} = \sigma(Z_i^T Z_j)$, where $\sigma$ represents the sigmoid function and $Z_i=f(v_i)$ and $Z_j=f(v_j)$ are the node representations obtained from GraphRank. The experimental results for link prediction, evaluated using the area under the ROC curve (AUC) and average precision (AP), are presented in Table~\ref{tab:LP}. It is observed that generative graph SSL methods, except for GraphMAE, generally outperform contrastive graph SSL methods. This can be attributed to the fact that generative methods often utilize graph structure reconstruction as an objective, which aligns well with the goal of the link prediction task. On both metrics, GraphRank outperforms most of the baselines used for comparison except SeeGera, which validates the effectiveness of our method on the link prediction task. 

\vpara{Graph Classification.} We conducted experiments on six distinct graph classification datasets, adhering to the setup defined by GraphMAE. After running the experiments five times, we reported the average 10-fold cross-validation accuracy along with the standard deviation. Our results, as displayed in Table~\ref{tab:gc}, demonstrate the effectiveness of our proposed method against self-supervised baselines, as it outperforms them in the majority of the datasets. Remarkably, our method achieves results that are comparable to supervised baselines.
The outcomes indicate that our approach is capable of learning meaningful information at the graph level, which subsequently proves advantageous for graph classification tasks.

\begin{table*}[htb]
\caption{Experimental results of graph classification. The best results are marked bold, while the second-best results are underlined.}
\label{tab:gc}
\begin{threeparttable}
\begin{tabular}{llccccc}
\hline
          & IMDB\_B             & IMDB\_M             & PROTEINS            & COLLAB              & MUTAG               & REDDIT-B         \\ \hline
GIN       & 75.1±5.1            & 52.3±2.8            & 76.2±2.8            & 80.2±1.9            & 89.4±5.6            & 92.4±2.5         \\
DiffPool  & 72.6±3.9            & -                   & 75.1±3.5            & 78.9±2.3            & 85.0±10.3           & 92.1±2.6         \\ \hline
WL        & 72.30±3.44          & 46.95±0.46          & 72.92±0.56          & -                   & 80.72±3.00          & 68.82±0.41       \\
DGK       & 66.96±0.56          & 44.55±0.52          & 73.30±0.82          & -                   & 87.44±2.72          & 78.04±0.39       \\ \hline
GraphCL   & 71.14±0.44          & 48.58±0.67          & 74.39±0.45          & 71.36±1.15          & 86.80±1.34          & {\ul 89.53±0.84} \\
JOAO      & 70.21±3.08          & 49.20±0.77          & {\ul 74.55±0.41}    & 69.50±0.36          & 87.35±1.02          & 85.29±1.35       \\
GCC       & 72.0                & 49.4                & -                   & 78.9                & -                   & \textbf{89.8}    \\
MVGRL     & 74.20±0.70          & 51.20±0.50          & -                   & -                   & {\ul 89.70±1.10}    & 84.50±0.60       \\
InfoGCL   & 75.10±0.90          & 51.40±0.80          & -                   & 80.00±1.30          & \textbf{91.20±1.30} & -                \\
GraphMAE  & {\ul 75.52±0.66}    & {\ul 51.63±0.52}    & \textbf{75.30±0.39} & {\ul 80.32±0.46}    & 88.19±1.26          & 88.01±0.19       \\ \hline
GraphRank & \textbf{75.65±0.45} & \textbf{52.07±0.20} & 73.36±0.14          & \textbf{81.30±0.12} & 89.36±0.02          & 87.30±0.20       \\ \hline
\end{tabular}
\begin{tablenotes}
        \footnotesize
      \item[] \scriptsize "-"  denotes unreported results due to either unavailability of the code or out-of-memory.
    \end{tablenotes}
      \end{threeparttable}
\end{table*}

Based on the experimental results presented, we can confidently answer $\mathcal{RQ}1$ as follows: GraphRank is an effective SSL method for graph data which is capable of learning superior node representations facilitating various downstream tasks.% i.e., node classification and link prediction.

\subsection{Model Analysis}
To answer $\mathcal{RQ}2$, in this subsection we conduct experiments to analyze the training efficiency of the model and the sensitivity of the model to hyperparameters.

\vpara{Comparison of Training Efficiency.}
\begin{figure}
    \centering
    \includegraphics[width=1\linewidth]{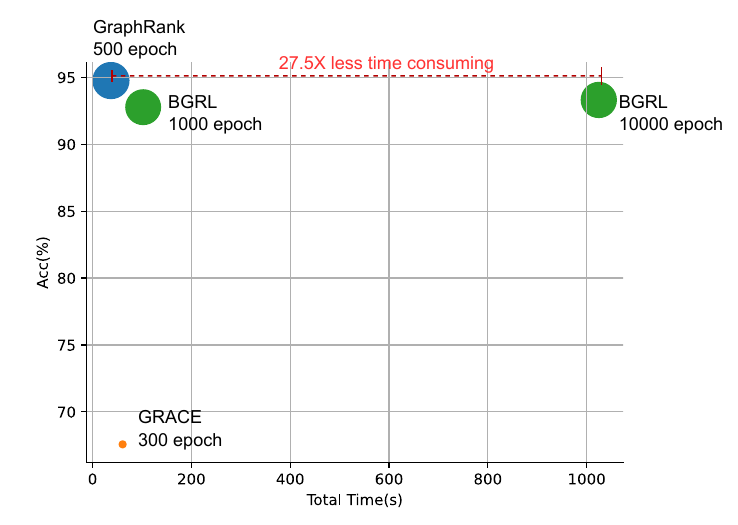}
    \caption{Comparison of training efficiency on the CS dataset.}
    \label{fig:time_cost}
\end{figure}
To evaluate training efficiency, we conduct experiments on the CS dataset comparing the performance of our proposed GraphRank with representative methods GRACE and BGRL. GRACE is a distinctive method that considers all nodes except the anchor nodes as negative samples, while BGRL is another well-known method that does not use negative samples.
As illustrated in Figure~\ref{fig:time_cost}, GraphRank greatly outperforms both GRACE and BGRL in terms of training time while achieving the best performance. Specifically, GraphRank takes 27.5 times less time than BGRL while achieving comparable performance. %Even if GRACE can converge with fewer epochs, its need for a large number of negative samples still results in it requiring 1.63 times more training time compared to GraphRank. Additionally, GRACE performs poorly on the CS dataset.
Even if GRACE can converge with fewer epochs, it still requires 1.63 times more training time than GraphRank due to its need for a large number of negative samples. Furthermore, GRACE performs poorly on the CS dataset.
We attempte to reduce the number of training epochs for BGRL to 1,000, which lead to a noticeable decline in its performance (Accuracy dropped from 93.3 to 92.7, whereas the Accuracy of GraphRank is 94.7). Even so, it still require 2.74 times the training time of GraphRank.
It is noteworthy that, despite BGRL exhibiting less computational expense within a single epoch compared to GraphRank, according to the official configurations provided by BGRL, it requires many more epochs (10,000 epochs) to achieve convergence. This results in an overall training time exceeding that of GraphRank.
Further experimental analysis on training efficiency can be found in Appendix~\ref{app:time}.

\begin{figure}[htb]
        \centering
        \includegraphics[width=0.45\textwidth]{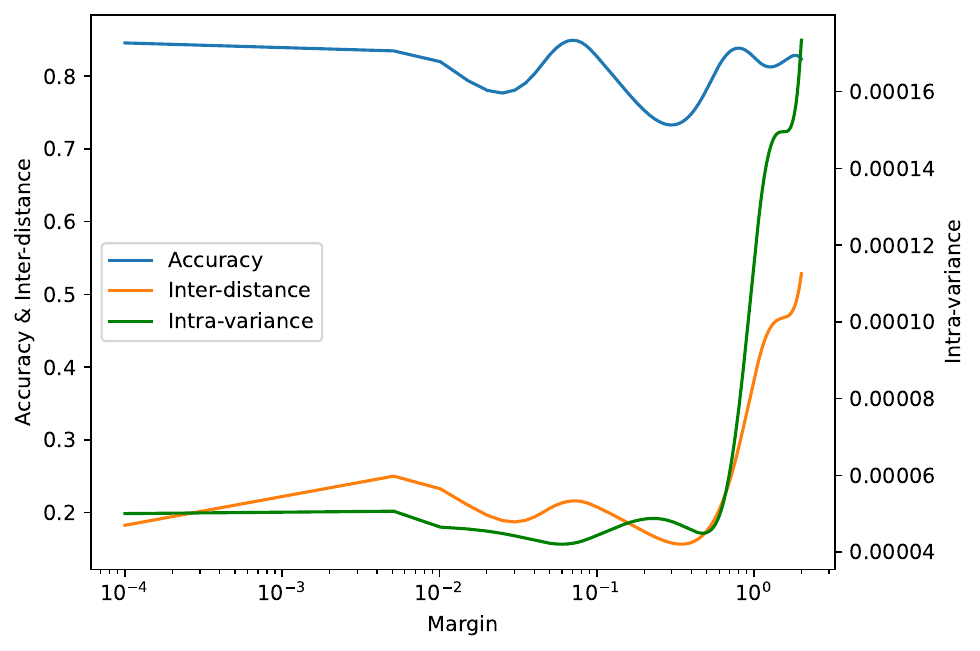}
        \caption{Impact of margin in Acc, Intra-class variance and Inter-class distance}
        \label{figure_impact_of_margin}
    \end{figure}

\vpara{Sensitivity Analysis of Margin.} For analyzing the effect of margin in Equation~\ref{equation:rank_loss} on the model, we conduct experiments on the Cora dataset with the margin ranging from 0 to 10 to evaluate its impact on the model's performance. The experimental results are presented in Figure \ref{figure_impact_of_margin}. We can observe that both the intra-class variance and inter-class distance increase as the margin value increases. The margin hyperparameter determines the threshold that distinguishes the similarity between positive pair from the similarity between negative pair. Although both intra-class variance and inter-class distance increase with the enlargement of the margin, the magnitude of inter-class distance is larger than that of Intra-class variance. Thus, the increase in inter-class distance is more significant. However, there is a slight drop in Acc when margin is greater than 1, which may be attributed to the sharp increase in Intra-class variance.

Now we can answer $\mathcal{RQ}2$: the training efficiency of GraphRank is obviously much better compared to GRACE. GraphRank requires considerably less training time compared to GRACE across multiple datasets, making it a more efficient choice for graph representation learning. When margin is smaller than a certain value, there is little impact on the model's performance, particularly in terms of node classification accuracy. This suggests that for practical training purposes, it is reasonable to set the margin value to a relatively small number, such as 0.0001, simplifying the training process without sacrificing performance.

\section{Conclusion}
    In this paper, we find two crucial flaws in existing methods using the InfoNCE loss: the false negative and high computational complexity. The false negative problem arises when intra-class nodes are mistakenly sampled as negative pairs,, and the model pushes them away, causing larger intra-class distance. And existing methods all require for lots of negative samples to work effectively, which results in high computational complexity, and make the models hard to be deployed on large graphs. To address these flaws, we propose GraphRank, by leveraging the rank loss, GraphRank mitigates the impact of false negatives and gather intra-class nodes closer; also GraphRank could perform satisfying with only few negative samples as it could solve the problem that intra-class negative nodes may be dominating when negative samples are small. Extensive experiments are conducted on GraphRank to evaluate its effectiveness and efficiency.
     
\newpage
\bibliographystyle{ACM-Reference-Format}
\bibliography{reference}

\appendix
\onecolumn
\section{Datasets} \label{data}

The first three datasets \cite{dataset_Cora_CiteSeer_PubMed2} are citation networks,where nodes represent individual papers, edges represent citation relationships between papers, and labels indicate the fields or topics of the papers. Photo and Computer \cite{dataset_Amazon_P_C} are derived from the Amazon product co-purchasing network, in these datasets, nodes represent the products and edges signify frequent co-purchasing relationships between products, labels represent the category of the product. Nodes in CS and Physics represent the author, the link stands for coauthor relationship between authors and labels are the fields of the authors. 

Our data split and experimental settings for node classification and link prediction tasks adhere to the established guidelines from SeeGera~\cite{SeeGera}. To be specific, in node classification tasks, three datasets - Cora, Citeseer, and Pubmed - each have 20 fixed nodes per class assigned for training, with an additional 500 nodes and 1000 nodes designated for validation and testing respectively. For the other four datasets, we randomly partition the nodes into 10\%/10\%/80\% subsets for train, validation, and test. In the link prediction tasks, we randomly divide the edges into 70\%/20\%/10\% subsets for train, validation, and test. This ensures a fair comparison with all baselines, as our approach is consistent with theirs. The data split employed in our experiments is the most prevalent strategy utilized on these classic datasets.

\begin{table}[!htbp]
      \caption{Statistics for node classification and link prediction}
      \label{Dataset_statistics}
      \centering
      \setlength{\tabcolsep}{1mm}{
        \begin{tabular}{ccccc}
            \toprule 
            Dataset & Nodes & Edges & Features & Classes \\
            \midrule
            Cora  & 2,708 & 5,429 & 1,433 & 7 \\
            Citeseer  & 3,327 & 4,732 & 3,703 & 6\\
            Pubmed  & 19,717 & 44,338 & 500 & 3\\
            Photo  & 7,650 & 119,081 & 745 & 8 \\
            Computers  & 13,752 & 245,861 & 767 & 10 \\
            CS &18,333 &182,121 &6,805 &15\\
            Physics &34,493 &530,417 &8,415 &5\\
            \bottomrule
        \end{tabular}
    }
    \end{table}

The statistical information of the graph classification datasets is presented in Table~\ref{tab:gc_data}. Each dataset comprises a collection of graphs, each associated with a label. The node labels are utilized as input features in MUTAG, PROTEINS, and NCI1, while node degrees are used as input features in IMDB-B, IMDB-M, REDDIT-B, and COLLAB.
    
% \usepackage{booktabs}
\begin{table}[!htbp]
\caption{Statistics for graph classification datasets.}
\label{tab:gc_data}
\begin{tabular}{@{}llllllll@{}}
\toprule
Dataset       & IMDB-B & IMDB-M & PROTEINS & COLLAB & MUTAG & REDDIT-B  \\ \midrule
 graphs     & 1,000   & 1,500   & 1,113     & 5,000   & 188   & 2,000    \\
 classes    & 2      & 3      & 2        & 3      & 2     & 2          \\
Avg. nodes & 19.8   & 13.0   & 39.1     & 74.5   & 17.9  & 429.7   \\ \bottomrule
\end{tabular}
\end{table}

These datasets are commonly used benchmarks in graph learning and analysis tasks, allowing for fair comparisons between different methods. Their diverse characteristics and structures enable the evaluation of GraphRank's performance across different domains.

\section{Additional Remarks} \label{app:remark}
\subsection{Intra-class Variance}
intra-class variance refers to the variance of node representations within each class. The specific calculation formula is as follows:
$$
\sigma^2=\frac{1}{\max (0, N_c-\delta N_c)} \sum_{i=0}^{N_c-1}\left(x_i-\bar{x}\right)^2,
$$
where $N_c$ is the number of nodes of class c and x denotes the node representation of class c obtained via GraphRank. $\delta N_c$ is the `correction`. In Figure~\ref{figure_intra-class_variance}, the intra-class variance represents the average variance across all classes.

\subsection{Inter-class Distance}
The Inter-class Distance is defined as the distance between classes, with the specific calculation formula as follows:
\begin{eqnarray*}
d=\frac{1}{\left | \mathcal{C} \right | (\left | \mathcal{C} \right |-1)}  \sum_{i \ne j,~  i~and~ j \in \mathcal{C} }^{} \left \| \bar{x}_i - \bar{x}_j \right \|,
\end{eqnarray*}
where  $\mathcal{C}$ represents the set of node classes, $i$ and $j$ specify the class labels, and $\bar{x}_i , \bar{x}_j$ are the average representations of nodes in each respective class.

\section{Additional Analysis}
\subsection{Comparison with InfoNCE}
In our opinion, contrastive learning with InfoNCE aims to separate samples from different class and gather nodes from the same class. However, the effectiveness of contrastive learning heavily relies on the availability of accurate negative samples. Due to data augmentation techniques, only the positive samples can be ensured to come from the same class, while negative samples may include samples from the same class (false negatives). This leads to suboptimal performance as the model mistakenly separates intra-class samples. The main challenge lies in defining negative samples, as contrastive learning requires nodes from different classes to be true negatives, which cannot be achieved during pre-training since class labels are not available.

    On the other hand, the rank loss in GraphRank alters the definition of negative samples. In GraphRank, positive similarities are required to be larger than negative similarities. This holds true regardless of whether the negative samples are from the same class or not, considering that positive samples are essentially the same sample. \textbf{Consequently, in GraphRank, the true negative samples are defined as all other nodes, rather than nodes from different classes.} As a result, GraphRank does not suffer from the false negative problem.

\subsection{Comparison with existing rank-based methods on graph}
Firstly, it is worth noting that there are currently few studies focusing on the application of rank loss within the field of graph self-supervised learning. As far as we know, previous works~\cite{ning2022graph,hoffmann2022ranking} use ranking to include more positive samples and determine the importance of positive pairs. But we aim to reduce the number of negative samples by leveraging rank loss. We point that contrastive learning requires lots of negative sample mainly because of false negative samples, and rank loss could relieve the false negative problem by not separating negative samples as far as possible, so we use rank loss only to ensure that positive pairs are more similar than negative ones, and the rank is between positive pair and negative pair while others use ranking between positive pairs. Therefore, we want to solve totally different problem, and use different methods.

\subsection{A small margin is enough}
We believe that there exists some false negative samples which are actually the same class of anchor node, so using a large margin would unavoidably push the intra-class nodes further apart while separating different classes. When operating with a small margin, the inter-class distance may already be sufficiently large to distinguish between inter-class nodes effectively. Increasing the margin further would not only expand the inter-class distance but also augment the intra-class variance, with the latter experiencing a more substantial increase as shown in Figure 6 from the paper. On the other hand, employing a small margin allows the model avoiding unnecessary separation of intra-class nodes while the inter-class ones are already separable. In nutshell, even with small margin, inter-class nodes would still be separable, and small margin could avoid intra-class separating.

\section{Additional Experiments} \label{exp}
\subsection{Node Clustering}
We have conducted additional experiments on node clustering. As illustrated in the Table~\ref{tab:cluster}, GraphRank achieves the best clustering results on 6 out of 7 datasets, indicating that GraphRank excels in obtaining superior node representations, which in turn benefits downstream tasks.
% \usepackage{booktabs}
\begin{table}[htb]
\caption{Experimental results for node clustering. The best results are marked bold.}
\label{tab:cluster}
\begin{tabular}{@{}lllllllllllllll@{}}
\toprule
Dataset   & \multicolumn{2}{c}{Cora}        & \multicolumn{2}{c}{Citeseer}    & \multicolumn{2}{c}{Pubmed}      & \multicolumn{2}{c}{Photo}       & \multicolumn{2}{c}{Computer}    & \multicolumn{2}{c}{CS}          & \multicolumn{2}{c}{Physics}     \\ \midrule
Metrics   & NMI            & ARI            & NMI            & ARI            & NMI            & ARI            & NMI            & ARI            & NMI            & ARI            & NMI            & ARI            & NMI            & ARI            \\ \midrule
GCN       & 47.42          & 46.55          & 27.74          & 28.65          & 33.94          & 38.67          & 81.73          & 83.99          & 70.99          & 72.60          & 88.18          & 91.36          & 86.68          & 92.64          \\
GAT       & 39.05          & 38.74          & 22.83          & 21.31          & 26.51          & 30.74          & 79.85          & 82.13          & 68.46          & 70.15          & 87.42          & 90.87          & 85.85          & 92.01          \\
GraphMAE  & \textbf{65.17} & 66.59          & 43.42          & 45.81          & 42.00          & 48.19          & 82.90          & 85.09          & 76.61          & 78.06          & 85.86          & 85.86          & 84.09          & 90.67          \\
SeeGera   & 62.92          & 63.76          & 43.86          & 45.44          & 39.63          & 45.35          & 76.01          & 77.05          & 35.44          & 39.79          & 87.41          & 90.70          & -              & -              \\ \midrule
GraphRank & 64.98          & \textbf{66.67} & \textbf{44.85} & \textbf{47.11} & \textbf{42.11} & \textbf{48.56} & \textbf{89.14} & \textbf{86.41} & \textbf{78.29} & \textbf{79.17} & \textbf{88.75} & \textbf{91.71} & \textbf{87.27} & \textbf{93.04} \\ \bottomrule
\end{tabular}
\end{table}

\subsection{Training Efficiency} \label{app:time}
% \paragraph{\textbf{Comparison of Training Efficiency}}

We conduct experiments on seven datasets to analyze the training efficiency of GraphRank and GRACE, a representative method of graph contrastive learning. To ensure fairness and consistency in the comparison, we keep the training parameters of GraphRank and GRACE the same including the learning rate, the number of GNN layers, the hidden layer dimension, etc., and the training epoch is fixed at 1000. As illustrated in Table~\ref{tab:time}, GraphRank spends significantly less time for training than GRACE on all seven datasets, specifically, GraphRank exhibits training efficiency that is 1.69 to 7.4 times higher than that of GRACE. This is consistent with our analysis in the previous subsection~\ref{sec:complexity}, since GraphRank requires only one negative sample for a target node, while GRACE treats all other nodes as negative samples. Moreover, the results in Table~\ref{tab:time} indicate that the efficiency advantage of GraphRank is especially prominent as the dataset size increases, indicating its scalability and practical utility in real-world scenarios. Additionally, we provide further analysis on convergence time in the Appendix~\ref{app:time}.

\begin{table}[htb]
\caption{The training time consumption comparison on seven
datasets.}
\label{tab:time}
\resizebox{\linewidth}{!}{
\begin{tabular}{@{}llllllll@{}}
\toprule
                                 & { Cora}  & { Citeseer} & { Pubmed} & { Photo} & { Computer} & { CS}     & { Physics} \\ \midrule
{ GRACE}     & { 14.06} & { 19.66}    & { 314.25} & { 73.01} & { 127.19}   & { 203.02} & { 1230.02} \\
{ GraphRank} & { 8.30}  & { 11.43}    & { 30.96}  & { 30.65} & { 57.97}    & { 74.51}  & { 166.32}  \\
{ Speed Up}  & \color{blue}{ 1.69$\times$} & \color{blue}{ 1.72$\times$}    & \color{blue}{ 10.15$\times$} & \color{blue}{ 2.38$\times$} & \color{blue}{ 2.19$\times$}    & \color{blue}{ 2.72{$\times$}}  & \color{blue}{ 7.40$\times$}   \\ \bottomrule
\end{tabular}}
\end{table}

Intuitively, the optimization objectives of GraphRank and GRACE are distinct from each other. GRACE employs InfoNCE as its optimization objective, aiming to maximize the consistency between positive samples and minimize the consistency between negative samples. In contrast, GraphRank utilizes rank loss as its optimization objective, with the purpose of ensuring that the similarity between the target node and positive samples exceeds the similarity between the target node and negative samples. It is inconclusive to claim that GraphRank and GRACE have significant differences in training epochs. On the other hand, this can be empirically verified through experimentation~\ref{tab:conv_time}. 
% \usepackage{booktabs}
% \usepackage{multirow}
\begin{table}[htb]
\caption{The convergence epochs and time.}
\label{tab:conv_time}
\begin{tabular}{@{}lllllll@{}}
\toprule
Dataset                    &        & Cora & Citeseer & Pubmed & Computer & CS    \\ \midrule
\multirow{2}{*}{GRACE}     & Epochs & 200  & 150      & 600    & 600      & 300   \\
                           & Time   & 2.81 & 2.95     & 188.55 & 76.31    & 60.91 \\ \midrule
\multirow{2}{*}{GraphRank} & Epochs & 150  & 150      & 200    & 500      & 500   \\
                           & Time   & 1.25 & 1.71     & 6.19   & 28.99    & 37.26 \\ \bottomrule
\end{tabular}
\end{table}

\end{document}